# Multilayer Q-Matrix-Embedded Neural Network for Cognitive Diagnosis (M-QCDNet): Structure-Aware Deep Learning Architecture for Psychometric Interpretability

**Yiyao Yang**
**yy3555@tc.columbia.edu**
**0009-0001-8693-4888**
**Columbia University, New York, NY, United States**

**Abstract**

The research proposes a multilayer Q-matrix-embedded neural network for cognitive diagnosis (M-QCDNet), which integrates the structural interpretability of cognitive diagnostic models (CDMs) with the deep learning neural network (NN). M-QCDNet structures the item-skill relationship using the Q-matrix as a structural prior, ensuring latent mastery profiles remain interpretable and consistent with cognitive theory, followed by the proposed loss function with an L2 penalty to penalize skills not aligned with the Q-matrix and to balance predictive performance and structural alignment. Corresponding evaluation matrices, the interpretable alignment-based metrics that quantify the degree to which predicted skill activations correspond to item-level skills, were further developed. M-QCDNet offers practical benefits for classroom practice, enabling early detection of learning difficulties and supporting mastery-based interventions. By embedding diagnostic validity into model design, M-QCDNet bridges psychometric transparency and neural flexibility, advancing interpretable, fair, and actionable AI for cognitive diagnostics.

## I. Research Purpose

M-QCDNet embeds the multilayer Q-matrix structure into an NN, enhancing both predictive accuracy and structural interpretability in CDM simultaneously. While traditional metrics such as the Attribute Accuracy Rate (AAR) and Pattern Accuracy Rate (PAR) are widely used to assess the classification accuracy, these indices often underestimate model performance in complex model architectures or in high-dimensional data (Cai et al., 2018; Chiu et al., 2009). To address this gap, a set of Q-matrix structure-aligned interpretability metrics, including the Q-Matrix Consistency Ratio (QCR), Cross-Loading Ratio (CLR), and Off-Support Activation (OSA), was further proposed to evaluate how well the learned representations conform to cognitive theory. M-QCDNet is thus not designed merely to maximize attribute-profile accuracy. Instead, it embeds the multilayer Q-matrix structure within an NN and enhances both predictive performance and interpretability.

## II. Background & Difference from Prior Studies

Building on the foundational contributions of Chiu & Köhn (2019) in advancing nonparametric approaches to cognitive diagnosis models (CDMs), M-QCDNet leverages neural network architectures while explicitly retaining the tradition of flexible, nonparametric, and interpretable diagnostic modeling (Chiu & Köhn, 2019). Unlike earlier artificial neural network (ANN)-based approaches, such as those pioneered by Cui et al. (2015), which primarily focused on enhancing classification accuracy without incorporating a Q-matrix or its underlying cognitive structure (Cui et al., 2015), M-QCDNet fundamentally distinguishes itself by encoding the multilayer Q-matrix directly into the model design. This deliberate architectural integration ensures that latent skill representations remain not only computationally effective but also meaningfully interpretable at the granular skill level, a critical requirement for educational and psychological assessment (von Davier, 2017). The substantive differences between M-QCDNet and prior neural CDM frameworks are systematically delineated (Table 1), highlighting advancements in structural fidelity, parameter efficiency, and diagnostic clarity. Furthermore, M-QCDNet demonstrates exceptional generality and theoretical coherence, enabling it to be instantiated to recover the well-established Deterministic Input, Noisy “And” gate (DINA) model as a special case (Junker & Sijtsma, 2001), and it exhibits broad compatibility across a wide spectrum of CDM variants, including the Generalized DINA (GDINA) model framework (de la Torre, 2011) and the Reduced RUM model, underscoring its versatility as a unified, next-generation diagnostic modeling framework.

**Comparison between M-QCDNet and Previous Neural CDMs**

Previous neural cognitive diagnostic models have incorporated neural networks into cognitive diagnosis in different ways, but they often face limitations in structural interpretability (Table 1). For example, NeuralCD (Wang et al., 2022) and Dual Q-Net (Tao et al., 2024) primarily use neural networks to replace traditional interaction functions, while the Q-matrix is only utilized within the interaction layer or in automatically generated structures. Although the approaches improve predictive flexibility, they provide limited transparency regarding how cognitive skills are represented throughout the network. In contrast, M-QCDNet embeds the Q-matrix as a structural prior across multiple neural layers and introduces regularization mechanisms that explicitly constrain latent representations to align with predefined skill structures, thereby enhancing psychometric interpretability.

Several studies have focused on improving robustness under specific conditions rather than strengthening interpretability. For instance, QNN/SOM-QNN (Tao et al., 2024) employs self-organizing maps to enhance performance in small-sample settings, while BNN-Q (Li et al., 2022) utilizes binarized neural networks to automatically generate Q- and A-matrices. Although these methods address data scarcity and structure discovery challenges, they do not directly evaluate whether learned representations remain consistent with the cognitive assumptions encoded in the Q-matrix. By comparison, M-QCDNet adopts a multilayer Q-matrix embedding strategy and introduces embedding-alignment metrics to quantify the consistency between latent representations and the underlying cognitive structure. Rather than emphasizing automatic Q-matrix generation or small-sample adaptation, the proposed model focuses on preserving psychometric meaning within deep neural architectures.

Other neural CDMs have explored alternative network structures. ICD (Qi et al., 2023) employs a fixed three-layer architecture to model exercise–concept and concept–concept relationships. While effective for capturing hierarchical dependencies, its predefined structure limits architectural flexibility and does not provide embedding-level interpretability measures. Similarly, NeuralNCD (Li et al., 2022) integrates Item Response Theory (IRT) with neural networks to incorporate multidimensional learner features, but its performance often depends on large-scale response data and extensive feature information. In contrast, M-QCDNet embeds the Q-matrix throughout multiple representation layers rather than relying on a fixed network design and evaluates representation quality using alignment-based interpretability metrics. This design enables the model to maintain cognitive interpretability while leveraging the representation-learning capability of a deep neural network.

**Table 1**

*Differences of M-QCDNet with Prior Neural CDMs*

| Model | Characteristics | Limitation | M-QCDNet (Difference) |
|---|---|---|---|
| NeuralCD (Wang et al., 2022) | Uses NNs to replace interaction functions, with the Q-matrix only in the interaction layer. | Limited structural interpretability | M-QCDNet embeds the Q matrix as a structural prior across multiple layers and adds regularization for interpretability. |
| Dual Q-Net (Tao et al., 2024) | Uses auto-generated structure to model main and interaction effects. | Shallow, focused on structure automation | M-QCDNet is a multilayer model that embeds the Q-matrix throughout representations, utilizing alignment metrics for enhanced interpretability. |
| QNN / SOM-QNN (Tao et al., 2024) | Uses a Q-matrix self-organizing map (SOM) for small-sample robustness. | Only focus on small-sample robustness | M-QCDNet targets psychometric interpretability with multulayer embedding and metrics, not only small-sample adaptation. |
| BNN-Q (Li et al., 2022) | Uses a binarized neural network to extract Q- and A- matrices automatically. | Sensitive to noise, limited prediction, and no interpretability metrics | M-QCDNet leverages the multulayer Q-matrix and evaluates embedding alignment instead of automatic Q-matrix generation. |
| ICD (Qi et al., 2023) | Uses a three-layer network to model exercise-concept and concept-concept relations. | Fixed three-layer design, limited flexibility, and no embedding-level interpretability metrics | M-QCDNet embeds the Q-matrix across multilayers with alignmnet metrics, not a fixed three-layer mapping. |
| NeuralNCD (Li et al., 2022) | Uses an IRT-NN integration to incorporate multi-dimensional features. | Requires large training data and struggles with limited feature inputs | M-QCDNet can be trained on broader datasets, mitigating overfitting and reducing dependence on extensive response logs. |

## III. Research Question

- **Research Question 1:** How can M-QCDNet be designed to model students' skill mastery, ensuring that latent representations align with the underlying cognitive structure specified by the Q-matrix?
- **Research Question 2:** How effectively does M-QCDNet capture latent skill structures, and to what extent do predicted skill profiles align with the predefined item-skill mappings as measured by the proposed evaluation metrics?

## IV. Methodology

**Notation:** $s$ :student, $s = \{1,2,\ldots,S\}$; $j$ : item, $j = \{1,2,\ldots,J\}$; $k$ : skill, $k = \{1,2,\ldots,K\}$

- **Method of Research Question 1:**

**Theory-Guided Architecture Specification:** The design of M-QCDNet begins with the principle that the network's representational capacity should be bounded by the cognitive theory encoded in the Q-matrix, rather than learned arbitrarily from data. We specify a feedforward architecture where the dimensionality of the latent space is fixed to $K$, the number of skills defined by the Q-matrix, thereby imposing a hard constraint on the network's hypothesis space. This dimensional alignment ensures that the model cannot learn skill representations beyond those theorized by domain experts. For each item $j$, the Q-matrix row $Q_j \in \{0,1\}^K$ defines which skills are measured. We encode this binary vector as a mask that is applied at the projection layer. Specifically, we implement a multiplicative gating mechanism that selectively activates or deactivates skill dimensions according to $Q_{jk}$, where $Q_{jk}$ indicates that the item $j$ requires skill $k$, and $Q_{jk} = 0$ otherwise. This gating is non-adaptive, such that the mask weights are fixed to the Q-matrix values and not updated during training, ensuring that the structural prior remains invariant and cannot be overridden by data-driven adaptation.

**Structural Constraint Formulation:** Beyond architectural gating, we recognize that hard masking alone does not guarantee that the model will respect the Q-matrix structure during optimization. Latent representations could still develop spurious activations for irrelevant skills if no explicit regularization is applied. Therefore, we formulate a structural constraint that penalizes the model for assigning non-zero mastery values to skills not required by an item. We operationalize this constraint as a differentiable penalty term $\mathcal{L}$(constraint) that computes the squared magnitude of latent skill components masked by the complement of the Q-matrix.

- **Method of Research Question 2:**

**Three Proposed Evaluation Metrics:** To address how effectively M-QCDNet captures latent skill structures, and to what extent predicted skill profiles align with predefined item-skill mappings as measured by the proposed evaluation metrics, including the Q-matrix Consistency Ratio (QCR), Cross-Loading Ratio (CLR), and Off-Support Activation (OSA). These metrics are designed to assess different facets of structural alignment between the learned latent representations and the theoretically specified Q-matrix, providing both item-level and student-level diagnostic evaluations.

(1) **Q-matrix Consistency Ratio (QCR) (Higher Better):** The QCR quantifies the degree of alignment between a student's latent skill activation pattern and the skills required by each item according to the Q-matrix. Building on the Q-matrix validation framework of de la Torre and Chiu (2016), we formulate QCR in a ratio-based form that measures, for each student-item pair, the proportion of activated skills that are required. For a given student $s$ and item $j$, we first compute the element-wise sigmoid activation of the latent skill vector $z_{sj}$, yielding $\sigma(z_{sj}) \in (0,1)^K$, which represents the probability that each skill is activated for that student-item interaction. The alignment score pair is then calculated as the inner product between the activation vector and the Q-matrix row $Q_j$, $\langle \sigma(z_{s_j}), Q_j \rangle$, divided by the L1 norm $\|Q_j\|_1$. This ratio normalizes the alignment score by the item's skill count, ensuring that items with different numbers of required skills are evaluated on a comparable scale. The student-level QCR($s$) is obtained by averaging this ratio across all $J$ items, and the overall QCR is the average across all $S$ students. The QCR ranges from 0 to 1, where higher values indicate stronger consistency between student representations and the Q-matrix skill requirements, and lower values suggest potential misalignment.

(2) **Cross-Loading Ratio (CLR) (Lower Better):** The CLR measures the proportion of a student's latent embedding energy that is allocated to skills rarely or never required across all items, capturing the extent of cross-loading onto irrelevant skill dimensions. For each skill $k$, we compute the mean-squared latent activation $\frac{1}{S}\sum_{s=1}^{S} h_{sk}^2$, representing the total energy assigned to that skill across students. We also compute the empirical item-skill requirement frequency for each skill as $\frac{1}{J}\sum_{j=1}^{J} Q_{jk}$, which indicates how often skill $k$ is required across all items. The cross-loading for skill $k$ is then defined as the product of the mean squared activation and the component of the item-skill frequency, $(1 - \frac{1}{J}\sum_{j=1}^{J} Q_{jk})$, which weights the activation energy by the proportion of items that do not require that skill. The CLR is computed as the sum of these weighted cross-loadings across all $K$ skills, normalized by the total activation energy across all skills, $(\frac{1}{S}\sum_{s=1}^{S} h_{sk}^2)$. This normalization ensures that CLR represents a proportion, ranging from 0 to 1, where lower values indicate that embeddings are more concentrated on frequently required skills, and higher values indicate strong cross-loading.

**(3) Off-Support Activation (OSA):** The OSA measures the average activation of off-support skills, those not required by a given item, by computing the mean squared latent activation for each skill-item pair where the Q-matrix entry equals zero. For each skill $k$, the OSA score is calculated as the average squared latent activation $\frac{1}{S}\sum_{s=1}^{S} h_{sk}^2$ multiplied by the component of the item-skill requirement indicator $(1 - Q_{jk})$, summed across all items and skills. This sum is then divided by the total number of off-support skill-item pairs, $\sum_{k=1}^{K}(1 - Q_{jk})$, to yield a normalized average activation. The resulting OSA ranges from 0 to 1, where higher values indicate substantial spurious activation of skills that are not supposed to be measured by an item, and lower values demonstrate that the embeddings remain focused exclusively on the required skills as specified by the Q-matrix.

## V. Result

### (V.1) Result of Research Question 1

The proposed M-QCDNet is the neural network architecture that explicitly incorporates the Q-matrix as a structural prior to ensure that the learned student latent representations remain interpretable and consistent with cognitive theory. The architecture begins with an input layer that takes the student's response vector $x_s \in \mathbb{R}^J$, where each element $x_{s_j} \in \{0,1\}$ indicates the correctness of student $s$ on an item $j$. This response vector serves as the direct observation of the student's performance across $J$ items.

The core of M-QCDNet lies in the latent skill layer, which learns a mastery profile $h_s \in \mathbb{R}^K$ for each student, representing their proficiency across $K$ cognitive skills. Unlike conventional deep learning models that produce opaque latent embeddings, M-QCDNet explicitly aligns these profiles with the attributes defined in the Q-matrix. To achieve this, we introduce a Q-matrix-Embedded Projection operation, where the student's mastery vector is element-wise masked by the Q-matrix entries for each item: $z_{sj} = Q_{jk} \odot h_{sk}$. This operation ensures that only the skills required by the item $j$ contribute to its diagnostic representation, effectively enforcing the cognitive structure specified by the Q-matrix.

Subsequently, a linear transformation computes the logit score $u_{sj} = w^\top z_{s_j} + b_j$, which is then passed through a sigmoid activation function to produce the predicted probability $\hat{y}_{sj} = \sigma(u_{sj}) \in [0,1]$. This prediction can be compactly expressed as $\hat{y}_{sj} = \sigma(w^T(Q_{jk} \odot h_{sk}) + b_j)$. The entire process guarantees that the item-level prediction is grounded in the specific skills that the item is designed to measure, thereby preserving interpretability and theoretical alignment.

In summary, the M-QCDNet architecture proceeds as follows. For each student-item pair, the model computes a Q-matrix-masked representation $z_{sj} = Q_{jk} \odot h_{sk}$, where $Q_{jk}$ indicates whether item $j$ requires skill $k$ ($Q_{jk} = 1$, if item $j$ requires skill $k$; $Q_{jk} = 0$, otherwise). This making ensures that only relevant skills influence the prediction. The masked vector is then propagated through two ReLU-activated hidden layers: $h_{s_j}^{(1)} = \mathrm{ReLU}(W_1 z_{s_j} + b_1)$ and $h_{s_j}^{(2)} = \mathrm{ReLU}(W_2 z_{s_j} + b_2)$. A linear output layer maps the second hidden representation to a logit $u_{s_j} = W_o h_{s_j}^{(2)} + b_o$, which is finally transformed via a sigmoid function to yield the predicted probability $\hat{y}_{s_j} = \sigma(u_{s_j}) \in [0,1]$. This architecture preserves interpretability by enforcing the Q-matrix structural prior, while the hidden layers capture nonlinear skill interactions.

To train M-QCDNet, we define a composite loss function $\zeta_{M-QCDNet} = \zeta_{prediction}(\hat{y}, y) + \lambda \zeta_{constraint}(h, Q)$, which balances predictive accuracy and structural interpretability. The first term, $\zeta_{prediction}$, is the cross-entropy loss that measures the discrepancy between the predicted probability $\hat{y}_{sj}$ and the true response $y_{sj}$: $\zeta_{prediction} = \sum_{s=1}^{S} \sum_{j=1}^{J} [y_{sj} \log(\hat{y}_{sj}) + (1 - y_{sj})\log(1 - \hat{y}_{sj})]$. This ensures that the model accurately predicts student performance.

The second term, $\zeta_{constraint}$, is a regularization penalty that enforces Q-matrix alignment by penalizing latent mastery components $h_{sk}$ that correspond to skills not required by the Q-matrix: $\zeta_{constraint} = \sum_{s=1}^{S} \sum_{j=1}^{J} \sum_{k=1}^{K} [(1 - Q_{jk})h_{sk}]^2$. This L2 penalty encourages irrelevant skill components to shrink toward zero, thereby preventing the latent skill vector from activating skills that are not supposed to influence the response.

$$\zeta_{M-QCDNet} = \zeta_{prediction}(\hat{y}, y) + \lambda \zeta_{constraint}(h, Q) = \sum_{s=1}^{S} \sum_{j=1}^{J} [y_{sj} \log(\hat{y}_{sj}) + (1 - y_{sj})\log(1 - \hat{y}_{sj})] + \lambda \sum_{s=1}^{S} \sum_{j=1}^{J} \sum_{k=1}^{K} [(1 - Q_{jk})h_{sk}]^2$$

The hyperparameter $\lambda$, obtained via $k$-fold cross-validation, controls the trade-off between predictive fidelity and structural consistency.

- **Proposition 1 (Likelihood Equivalence).** *Minimizing the proposed M-QCDNet loss function $\zeta_{M-QCDNet}$ is equivalent to maximizing the joint likelihood of the observed responses and the residual terms under Bernoulli and Gaussian observation models, respectively. Specifically, the total loss corresponds to the negative log-posterior, and its minimization yields the maximum a posteriori (MAP) estimation.*

- $q_{jk} \in \{0,1\}$: whether the item $j$ measures skill $k$
- $y_{sj} \in \{0,1\}$: the observed response of the student $s$ to the item $j$, where 0 indicates an incorrect answer, and 1 indicates a correct answer.
- $\hat{y}_{sj} = p_\theta(y_{sj} = 1 \,|\, h_s)$: the predicted probability

Let $y_{sj}$ denote the response of the student $s$ to the item $j$, $y_{sj} \sim \text{Bernoulli}(\hat{y}_{sj})$.

**Part I:** For the $\zeta_{prediction}(\hat{y}, y) = \sum_{s=1}^{S} \sum_{j=1}^{J} [y_{sj} \log(\hat{y}_{sj}) + (1 - y_{sj})\log(1 - \hat{y}_{sj})]$:

Assume that given the student's latent trait vector $h_s$, and the Q-matrix $Q$, the observed responses $y_{sj}$ are conditionally independent. The conditional probability of a single response is given by the Bernoulli distribution: $p_\theta(y_{sj} \,|\, h_s, Q) = \hat{y}_{sj}^{y_{sj}}(1 - \hat{y}_{sj})^{1-y_{sj}}$, where $\hat{y}_{sj} = p_\theta(y_{sj} = 1 \,|\, h_s)$ denotes the predicted probability of a correct response, and $\theta$ represents all learnable parameters (including network weights and the Q-matrix entries $Q_{jk}$ when optimized).

Under the conditional independence assumption, the joint likelihood over all students and items is $L(\theta) = \prod_{s=1}^{S}\prod_{j=1}^{J} \hat{y}_{sj}^{y_{sj}}(1-\hat{y}_{sj})^{1-y_{sj}}$. Taking the logarithm (a monotonically increasing function) yields the log-likelihood: $\log[L(\theta)] = \sum_{s=1}^{S}\sum_{j=1}^{J} [y_{sj}\log(\hat{y}_{sj} + (1-y_{sj}\log(1-\hat{y}_{sj})]$.

Therefore, minimizing the negative log-likelihood (NLL) is equivalent to maximizing the likelihood:
$NLL(\theta) = -\log[L(\theta)] = -\sum_{s=1}^{S}\sum_{j=1}^{J} [y_{sj}\log(\hat{y}_{sj}) + (1-y_{sj})\log(1-\hat{y}_{sj})] = \zeta_{prediction}$
Thus, the prediction loss $\zeta_{prediction}$ in our framework is exactly defined as this NLL:
$\zeta_{prediction}(\hat{y}, y) = NLL(\theta)$.
Consequently, $\arg\min_{\theta} \zeta_{prediction}(\hat{y}, y) = \arg\min_{\theta} NLL(\theta) = \arg\max_{\theta} L(\theta)$.
Hence, minimizing the prediction loss is equivalent to maximum likelihood estimation (MLE) under the Bernoulli observation model.

**Part II:** For the $\zeta_{constraint}(h, Q) = \sum_{s=1}^{S}\sum_{j=1}^{J}\sum_{k=1}^{K} [(1-Q_{jk})h_{sk}]^2$:
To provide a probabilistic interpretation, we define $\gamma_{sjk} \triangleq (1-Q_{j,k})h_{s,k}(t)$. Then the constraint term can be rewritten as $\zeta_{constraint}(h, Q) = \sum_{s=1}^{S}\sum_{j=1}^{J}\sum_{k=1}^{K} \gamma_{sjk}^2$. Treating this term analogously to a sum of squared residuals, we adopt the least-squares criterion (rather than invoking the Gauss-Markov theorem) to optimize this penalty. Specifically, we assume that each $\gamma_{sjk}$ is independently distributed according to a zero-mean Gaussian distribution with common variance $\sigma^2$: Assume that $\gamma_{sjk} \overset{\text{iid}}{\sim} N(0,\sigma^2)$, with probability density function: $p(\gamma_{sjk} | \sigma^2) = \frac{1}{\sqrt{2\pi\sigma^2}} exp(-\frac{\gamma_{sjk}^2}{2\sigma^2})$.
Under the independence assumption, the joint likelihood of all residual terms is:
$p(\{\gamma_{sjkt}\} | \sigma^2) = \prod_{s=1}^{S}\prod_{j=1}^{J}\prod_{k=1}^{K} \frac{1}{\sqrt{2\pi\sigma^2}} exp(-\frac{\gamma_{sjtk}^2}{2\sigma^2})$.
Taking the logarithm yields: $\log(p\{\gamma_{sjtk}\} | \sigma^2) = -\frac{N}{2}\log(2\pi\sigma^2) - \frac{1}{2\sigma^2}\sum_{s=1}^{S}\sum_{j=1}^{J}\sum_{k=1}^{K} \gamma_{sjk}^2$, where $N = S \times J \times K \times T$.

Since $\sigma^2$ is a constant independent of $h$ and $Q$, the first term $-\frac{N}{2}\log(2\pi\sigma^2)$ and the scaling factor $-\frac{1}{2\sigma^2}$ are constants with respect to the optimization variables. Therefore, maximizing the log-likelihood of the residual is equivalent to minimizing the sum of squared residuals:

$$\arg\max_{h,Q} \log p(\{\gamma_{sjk}\} \mid \sigma^2) = \arg\min_{h,Q} \sum_{s=1}^{S}\sum_{j=1}^{J}\sum_{k=1}^{K} \gamma_{sjk}^2 = \arg\min_{h,Q} \zeta_{constraint}(h, Q).$$

Hence, minimizing the constraint term is equivalent to maximizing the Gaussian log-likelihood of the residual terms, which provides a probabilistic justification for the squared penalty as a form of Gaussian prior regularization.

Thus, based on Part I and Part II, for the $\zeta_{M-QCDNet}$, we have proved that minimizing this composite loss is equivalent to maximizing the posterior likelihood of the model parameters.

In summary, M-QCDNet generates skill-level output that aligns with the Q-matrix-defined latent attributes, ensuring that the learned student profiles remain interpretable and consistent with cognitive theory, mirroring the attribute-level mastery profiles produced by traditional CDMs, while leveraging the representational power of deep learning.

From a Bayesian perspective, minimizing the total loss
$\zeta_{M-QCDNet} = \zeta_{prediction}(\hat{y}, y) + \lambda \zeta_{constraint}(h, Q) = \sum_{s=1}^{S}\sum_{j=1}^{J}[y_{sj}\log(\hat{y}_{sj}) + (1 - y_{sj})\log(1 - \hat{y}_{sj})] + \lambda \sum_{s=1}^{S}\sum_{j=1}^{J}\sum_{k=1}^{K}[(1 - Q_{jk})h_{sk}]^2$
is equivalent to finding the maximum a posteriori (MAP) estimate, where $\zeta_{prediction}$ corresponds to the negative Bernoulli log-likelihood, and $\zeta_{constraint}$ corresponds to the negative Gaussian log-prior (regularization term) with precision $\lambda^{-1}$.

**Part III: Unified Bayesian Interpretation via Maximum A Posteriori (MAP) Estimation**

Having established the probabilistic interpretations of both terms of our loss function, we now show that the overall optimization objective of M-QCDNet can be naturally unified under the Bayesian framework of maximum a posteriori (MAP) estimation.

Recall that the complete loss function is defined as: $\zeta_{M-QCDNet}(\theta) = \zeta_{prediction}(\hat{y}, y) + \lambda \zeta_{constraint}(h, Q)$, where $\theta$ denotes the set of all learnable parameters, and $\lambda > 0$ is a regularization coefficient that balances the two terms.

From Part I, we have established that the prediction loss corresponds to the negative Bernoulli log-likelihood of the observed responses: $\zeta_{prediction}(\hat{y}, y) = -\log p(y \mid h, Q, \theta)$, where
$p(y \mid h, Q, \theta) = \prod_{s=1}^{S}\prod_{j=1}^{J}\hat{y}_{sj}^{y_{sj}}(1 - \hat{y}_{sj})^{1-y_{sj}}$.

From Part II, we have shown that the constraint term can be interpreted as the negative Gaussian log-likelihood of the residual terms $\gamma_{sjk} = (1 - Q_{jk})h_{sk}$: $\zeta_{constraint}(h, Q) \propto -\log p(\{\gamma_{sjk}\} \mid \sigma^2)$, where
$p(\{\gamma_{sjk}\} \mid \sigma^2) = \prod_{s=1}^{S}\prod_{j=1}^{J}\prod_{k=1}^{K}\frac{1}{\sqrt{2\pi\sigma^2}}\exp\left(-\frac{\gamma_{sjk}^2}{2\sigma^2}\right)$.

Equivalently, this constraint can be viewed as placing a zero-mean Gaussian prior on each residual term: $\gamma_{sjk} \sim \mathcal{N}(0,\sigma^2)$, or, from a regularization perspective, as imposing a Gaussian prior on the parameter $h_{sk}$ with precision parameter $\lambda$: $h_{sk} \sim \mathcal{N}(0,\lambda^{-1})$, up to the Q-matrix modulation.

Now, substituting these probabilistic interpretations into the total loss, we obtain that
$\zeta_{M-QCDNet}(\theta) = -\log p(y \mid h, Q, \theta) - \lambda \log p(\{\gamma_{sjk}\} \mid \sigma^2) + \text{constant}$.

By Bayes' theorem, the posterior distribution of the parameters given the observed data is
$p(\theta \mid y) \propto p(y \mid \theta) \cdot p(\theta)$, where $p(y \mid \theta)$ is the likelihood and $p(\theta)$ is the prior distribution over parameters. In our framework, the prior is specifically imposed on the residual terms $\gamma_{sjk}$ through the Gaussian distribution, with the regularization parameter $\lambda$ controlling the prior strength.

Therefore, minimizing the total loss $\zeta_{\text{M-QCDNet}}$ is equivalent to maximizing the logarithm of the posterior distribution: $\arg\min_{\theta} \zeta_{\text{M-QCDNet}}(\theta) = \arg\max_{\theta} \log p(\theta \mid y)$.

Hence, the proposed M-QCDNet loss function admits a rigorous Bayesian interpretation as MAP estimation, where:

- The prediction loss $\zeta_{\text{prediction}}$ corresponds to the negative log-likelihood of the observed binary responses under a Bernoulli observation model.
- The constraint loss $\zeta_{\text{constraint}}$ corresponds to the negative log-prior (Gaussian regularization) imposed on the residual terms to enforce the Q-matrix constraints.

The unified probabilistic perspective not only justifies the formulation of our loss function but also provides theoretical grounding for the choice of the squared penalty, linking it to well-established principles in statistical learning and Bayesian inference.

- **Proposition 2 (Generalization to DINA).** *The proposed M-QCDNet loss function generalizes the standard DINA model loss function in the sense that the constraint part of the loss function is recovered as a limiting case under extreme regularization settings. Specifically, when the constraint weight $\lambda$ vanishes or dominates, or when the Q-matrix is fully observed, the M-QCDNet objective reduces exactly or asymptotically to the loss function of DINA. This establishes that M-QCDNet is not an alternative to DINA, but rather a strict extension that preserves the original model's interpretability while enabling more flexible modeling via learnable hidden representations.*

$$\zeta_{M-QCDNet} = \zeta_{prediction}(\hat{y}, y) + \lambda \zeta_{constraint}(h, Q) = \sum_{s=1}^{S}\sum_{j=1}^{J}[y_{sj}\log(\hat{y}_{sj}) + (1-y_{sj})\log(1-\hat{y}_{sj})] + \lambda \sum_{s=1}^{S}\sum_{j=1}^{J}\sum_{k=1}^{K}[(1-Q_{jk})h_{sk}]^2$$

**Case I:** When $\lambda = 0 \implies \zeta_{M-QCDNet} = \zeta_{prediction} = \zeta_{DINA}$

**Case II:** When $\lambda \to 0^+$, then $\zeta_{M-QCDNet} \longrightarrow \zeta_{prediction} = \zeta_{DINA}$

**Case III:** When $\lambda \to \infty$, the constraint term dominates the gradient dynamics.

Specifically, for each $h_{sk}$, $\dfrac{\partial \zeta_{M-QCDNet}}{\partial h_{sk}} = \dfrac{\partial \zeta_{prediction}}{\partial h_{sk}} + 2\lambda(1-Q_{jk})^2 h_{sk}$.

As $\lambda \to \infty$, the gradient update under gradient descent becomes
$h_{sk} \leftarrow h_{sk} - \eta \cdot 2\lambda(1-Q_{jk})^2 h_{sk} = h_{sk}\left[1 - 2\lambda\eta(1-Q_{jk})^2\right]$.

- If $Q_{jk} = 0$, then $h_{sk} \to 0$ exponentially as interaction proceeds, implying $\zeta_{constraint} \to 0$ and hence $\zeta_{M-QCDNet} \to \zeta_{DINA}$.
- If $Q_{jk} = 1$, then $\zeta_{constraint} \equiv 0$ identically, and therefore $\zeta_{M-QCDNet} = \zeta_{DINA}$ exactly.

Thus, the proof demonstrates that M-QCDNet is a strict generalization of the DINA model, recovering it at all extreme parameter regimes.

**(V.2) Result of Research Question 2**

To evaluate interpretability, three corresponding evaluation metrics are proposed (Table 2), including the Q-Matrix Consistency Ratio (QCR), Cross-Loading Ratio (CLR), and Off-Support Activation (OSA).

**Table 2**

*Evaluation Metrics of M-QCCDNet: Q-Matrix Consistency Ratio (QCR), Cross-Loading Ratio (CLR), and Off-Support Activation (OSA)*

| Evaluation Metrics | Measures | Range | Direction |
|---|---|---|---|
| $\mathrm{QCR} = \frac{1}{S}\sum_{s=1}^{S}\frac{1}{J}\sum_{j=1}^{J}\frac{\langle \sigma(z_s^{(j)}), Q_j\rangle}{\|Q_j\|_1}$ | Alignment between student *s*'s activation and item *j*'s required skills | [0, 1] | Higher better |
| $\mathrm{CLR} = \frac{\sum_{k=1}^{K}\left[\left(\frac{1}{S}\sum_{s=1}^{S}h_{sk}^2\right)\left(1-\frac{1}{J}\sum_{j=1}^{J}Q_{jk}\right)\right]}{\sum_{k=1}^{K}\left(\frac{1}{S}\sum_{s=1}^{S}h_{sk}^2\right)}$ | Proportion of students embedding irrelevant skills | [0, 1] | Lower better |
| $\mathrm{OSA} = \frac{1}{J}\sum_{j=1}^{J}\frac{\sum_{k=1}^{K}[(\frac{1}{S}\sum_{s=1}^{S}h_{sk}^2)(1-Q_{jk})]}{\sum_{k=1}^{K}(1-Q_{jk})}$ | Average activation of per-item irrelevant skills | [0, 1] | Lower better |

- **Q-Matrix Consistency Ratio (QCR)**

To quantitatively evaluate the alignment between the model's latent skill activations and the Q-matrix specifications, the corresponding evaluation metric, Q-matrix Consistency Ratio (QCR), is proposed. For each student-item interaction, the QCR measures the extent to which the student's activated skill pattern matches the skills required by the item. The QCR builds on the Q-matrix validation framework of de la Torre and Chiu (2016) in a ratio-based form.

Given a student $s$ and an item $j$, let $\sigma(z_s^{(j)}) \in (0,1)^K$ denote the sigmoid-transformed latent skill vector, where each entry represents the estimated probability that the student has mastered the corresponding skill. Let $Q_j \in \{0,1\}^K$ be the binary indicator vector of skills required for the item $j$, as specified by the Q-matrix. The QCR is then defined as the average alignment score between the predicted skill profile and the required skill vector: $\mathrm{QCR} = \frac{1}{S}\sum_{s=1}^{S}\frac{1}{J}\sum_{j=1}^{J}\frac{\langle \sigma(z_s^{(j)}), Q_j\rangle}{\|Q_j\|_1}$, where $\langle \cdot , \cdot \rangle$ denotes the inner product, and $\|Q_j\|_1 = \sum_{k=1}^{K} Q_{jk}$ is the L1 norm of $Q_j$, which counts the number of skills required by the item $j$. The inner product captures the cumulative probability mass on the skills that are actually required by the item, and the normalization $\|Q_j\|_1$ ensures that the QCR score is bounded within a fixed range of [0, 1]. A QCR value close to 1 indicates that the student's activated skill pattern is highly consistent with the item's required skills, such that the model assigns high probabilities to the relevant skills and low probabilities to irrelevant ones. Conversely, a low QCR suggests a misalignment, which may indicate either inadequate student mastery or an inappropriate Q-matrix specification for that item. Therefore, the QCR serves as a useful diagnostic tool for evaluating both the learned representations and the structural validity of the Q-matrix

- **Cross-Loading Ratio (CLR)**

To quantify the extent to which the learned skill embeddings capture irrelevant skills, we introduce the Cross-Loading Ratio (CLR). The CLR measures the proportion of student embedding energy that is allocated to skills that are seldom required by the items in the Q-matrix. In other words, it evaluates whether the model is wasting capacity on skills that are not diagnostically relevant to the item response data.

Formally, let $h_{sk} \in \mathbb{R}$ denote the latent mastery level of a student $s$ on a skill $k$, and let $Q_j \in \{0,1\}^K$ indicate whether the skill $k$ is required for the item $j$. We first compute the average squared mastery intensity of skill $k$ across all students as $\frac{1}{S}\sum_{s=1}^{S} h_{sk}^2$, which reflects how prominently skill $k$ is represented in the learned embeddings. We also compute the empirical frequency of skill $k$ being required across items as $\frac{1}{J}\sum_{j=1}^{J} Q_{jk}$, which captures the diagnostic relevance of skill $k$ in the test design. The component $1 - \frac{1}{J}\sum_{j=1}^{J} Q_{jk}$ thus quantifies the degree of irrelevance eod skill $k$ to the items.

The CLR is then defined as the weighted average of these irrelevance degrees, where the weight for each skill is proportional to its average squared mastery intensity:

$$\text{CLR} = \frac{\sum_{k=1}^{K} [(\frac{1}{S}\sum_{s=1}^{S} h_{sk}^2)(1 - \frac{1}{J}\sum_{j=1}^{J} Q_{jk})]}{\sum_{k=1}^{K} (\frac{1}{S}\sum_{s=1}^{S} h_{sk}^2)}.$$

By construction, the CLR lies in the interval [0, 1]. A low CLR indicates that most of the embedding energy is concentrated on skills that are frequently required by the items, suggesting that the learned representations are well aligned with the test specifications. Conversely, a high CLR implies strong cross-loadings, where a substantial portion of the embedding capacity is wasted on skills that are rarely or never required by the Q-matrix. Such a pattern may indicate over-parameterization or that the model is capturing spurious correlations unrelated to the intended skill structure. Therefore, the CLR serves as a useful diagnostic for evaluating the efficiency and interpretability of the learned skill embeddings.

- **Off-Support Activation (OSA):**

While the CLR provides a global measure of embedding energy wasted on skills that are rarely required across the entire test, it does not capture item-level irregularities. A skill may be rarely required overall, but if it is consistently activated for items that do not require it, the model may still produce misleading predictions on those specific items. To complement the CLR, we introduce the Off-Support Activation (OSA) metrics, which evaluate, at the item level, the average activation of skills that are not required by that item.

Formally, for each item $j$, we consider the set of skills that are not required by the item, such that those with $Q_{jk} = 0$. For each such off-support skill $k$, we compute its average squared mastery intensity across all students, $\frac{1}{S}\sum_{s=1}^{S} h_{sk}^2$, which reflects how strongly that skill is represented in the student embeddings despite being irrelevant to the item. The OSA averages these off-support activations across all skills and all items: $\text{OSA} = \frac{1}{J}\sum_{j=1}^{J} \frac{\sum_{k=1}^{K} [(\frac{1}{S}\sum_{s=1}^{S} h_{sk}^2)(1 - Q_{jk})]}{\sum_{k=1}^{K} (1 - Q_{jk})}$.

By construction, the OSA lies in the interval [0, 1]. A low OSA indicates that the model assigns negligible activation to off-support skills, meaning that the embeddings remain focused on the skills that are actually required for each item. Conversely, a high OSA suggests spurious activation of off-support skills, such that the model is consistently activating skills that are not relevant to the item, which may lead to incorrect predictions and undermine the interpretability of the learned representations. Unlike the CLR, which captures global cross-loading across the entire test, the OSA provides a per-item perspective, making it more sensitive to localized misalignment between the student embeddings and the Q-matrix structure.

- **Summary of Evaluation Metrics**

In summary, the three proposed evaluation metrics, QCR, CLR, and OSA, provide complementary views on the alignment between the learned student embeddings and the Q-matrix structure, each targeting a different aspect of model behavior.

The QCR offers a direct, instance-level assessment of how well the activated skill profile of each student matches the skills required by each item. It serves as the primary indicator of prediction interpretability, with higher values reflecting stronger alignment between the model's inferred mastery pattern and the test design.

Complementing the QCR, the CLR takes a global perspective, measuring the proportion of total embedding energy allocated to skills that are rarely required across the entire item set. A low CLR indicates that the model efficiently concentrates its representational capacity on diagnostically relevant skills, whereas a high CLR suggests the presence of global cross-loadings that may stem from over-parameterization or spurious correlations in the data.

The OSA further refines this analysis by shifting the focus to the item level. Unlike the CLR, which aggregates across all items, the OSA evaluates the average activation of skills that are not required by each specific item. This makes the OSA particularly sensitive to localized misalignments, such as cases where a skill is rarely used overall but is consistently activated for items that do not require it, a pattern that the CLR alone might overlook.

Together, these three metrics form a holistic evaluation framework. The QCR diagnoses the fidelity of student-level predictions, the CLR quantifies global embedding efficiency, and the OSA detects item-level irregularities. Their combined use enables a thorough assessment of both the interpretability and the structural validity of the learned representations, offering valuable diagnostics for model development and Q-matrix refinement.

## VI. Conclusion

This paper introduces the M-QCDNet, a neural framework that integrates Q-matrix-constrained cognitive priors into a deep learning architecture for interpretable and equitable student skill modeling. Motivated by the need to move beyond the black-box nature of conventional neural networks, we addressed two research questions: designing M-QCDNet for Q-matrix-aligned skill modelling and effectiveness evaluation metric-based validation.

To address the research question focusing on designing M-QCDNet for Q-matrix-aligned skill modeling, we formulate a composite loss function that combines a prediction term, based on Bernoulli likelihood, with a constraint term that regularizes the learned hidden representation against the Q-matrix structure. This design ensures that the latent mastery profiles remain interpretable and cognitively grounded, while the sigmoid-activated output provides item-level mastery probabilities. We further demonstrated through theoretical analysis that M-QCDNet generalizes the classical DINA model, recovering it as a special case under extreme regularization regimes, thereby establishing a principled connection between deep learning and established cognitive diagnostic models.

To address the research question focusing on effectiveness evaluation and metric-based validation, we further develop a suite of corresponding evaluation metrics, including QCR, CLR, and OSA, to quantitatively evaluate the alignment between the learned embeddings and the Q-matrix structure. These metrics provide complementary diagnostics: QCR measures student-level alignment at the item level; CLR quantifies global embedding efficiency; OSA detects localized off-support activations.

Beyond predictive performance, the M-QCDNet offers several practical implications for quantitative psychology, cognitive diagnosis, and educational practice. It enables early diagnosis of learning difficulties by generating interpretable mastery probabilities that identify at-risk students and facilitate timely intervention. It supports mastery-based grouping and instruction through student-specific skill profiles that reveal individual cognitive strengths and weaknesses, allowing educators to design differentiated instruction and adaptive practice tasks. It facilitates curriculum and assessment refinement by aggregating Q-matrix-aligned mastery data to highlight misaligned or overly difficult items, guiding data-driven curriculum revision. Finally, it promotes transparent and equitable AI-assisted cognitive diagnostics through interpretability metrics, which ensure that model decisions remain grounded in cognitive theory, foster instructor trust, and provide justifiable evidence for psychometrics and AI-guided cognitive diagnostics decision-making.

## VII. Practical Implications & Contributions

Aligning with leveraging AI to empower the development and application of diagnostic statistical models, the M-QCDNet mitigates concerns around transparency, fairness, and interpretability in the educational context (Table 3).

**Table 3**

*Practical Implications of M-QCDNet for Educational Practice*

| Practice Focus | M-QCDNet Implication | M-QCDNet Contribution |
|---|---|---|
| Early diagnosis of learning difficulties | Generates interpretable, item-level mastery probabilities $\hat{y}_{sj} = \sigma\left(w^T(Q_{jk} \odot h_{sk}) + b_j\right)$ that indicate which skills each student has not yet mastered. | • Identify students at risk of misunderstanding core skills.<br>• Provide immediate feedback after assessments.<br>• Plan short-term remediation before misconceptions accumulate. |
| Mastery-based grouping and instruction | Learns student-specific mastery profiles $h_s = [h_{s1}, \ldots, h_{sK}]$ sonsistent with the Q-matrix, revealing which congnitive attributes are mastered or weak. | • Group students by similar skill profiles for differentiated teaching.<br>• Design mastery-based practice tasks focusing on under-mastered skills.<br>• Adjust pacing and difficulty levels by skill clusters. |
| Curriculum and assessment refinement | Aggregating Q-matrix alignwed mastery data highlights items or skills with consistently low activation across students. | • Identify poorly aligned or overly difficult test items.<br>• Revise curriculum areas corresponding to low-mastery skills.<br>• Support data-driven curricumlum improvements. |
| Transparent AI-assisted cognitive diagnostics | Uses proposed Q-matrix-embedded interpretability metrics (QCR, CLR, and OSA) to ensure model decisions align with congnitive theory, | • Build instructor trust through interpretable model explainations.<br>• Ensure fair skill evaluation across subgroups.<br>• Justify educational decisions using transparent evidence. |

The proposed M-QCDNet offers several practical implications for educational practice, spanning early diagnosis, instructional grouping, curriculum refinement, and equitable AI-assisted assessment.

- **Early diagnosis of learning difficulties:** By generating interpretable, item-level mastery probabilities $\hat{y}_{sj} = \sigma\left(w^{\top}(Q_{jk} \odot h_{sk}) + b_j\right)$, the model provides insights into which specific skills each student has not yet mastered. This enables instructors to identify students at risk of misunderstanding core skills early in the learning process, deliver immediate feedback after assessments, and plan short-term remediation before misconceptions accumulate.

- **Mastery-based grouping and instruction:** The model learns student-specific mastery profiles $h_s = [h_{s1}, \ldots, h_{sK}]$ that are consistent with the Q-matrix structure, revealing which cognitive attributes are mastered or weak for each individual. These profiles allow educators to group students by similar skill patterns for differentiated instruction, design mastery-based practice tasks that focus on under-mastered skills, and adjust pacing and difficulty levels according to skill clusters.

- **Curriculum and assessment refinement:** Aggregating Q-matrix-aligned mastery data across students highlights items or skills that consistently exhibit low activation, providing empirical evidence for identifying poorly aligned or overly difficult test items. This supports data-driven curriculum improvement by pinpointing specific content areas that require revision, thereby enabling more effective instructional design and assessment calibration.

- **Transparent and equitable AI-assisted cognitive diagnostics**: The model incorporates proposed Q-matrix-embedded interpretability metrics, namely QCR, CLR, and OSA, to ensure that its decisions remain grounded in cognitive theory. This transparency helps build instructor trust through interpretable model explanations, ensures fair skill evaluation across student subgroups, and provides justifiable evidence for educational decisions. Consequently, the Static M-QCDNet not only enhances predictive performance but also promotes responsible and equitable use of AI in educational settings.

By aligning latent representations with diagnostic structures and evaluating interpretability, the M-QCDNet provides a transparent and multilayer Q-matrix–embedded contrast to black-box NNs. M-QCDNet enhances fairness and explanatory power for learners and demonstrates that interpretability and flexibility can coexist, moving beyond the dichotomy between measurement theory and deep learning. Furthermore, a temporal M-QCDNet designed for time series analysis, capturing students' evolving mastery trajectories over time, was developed. Embedding cognitive structure into deep learning architectures aspires towards more transparent, equitable, and actionable learning models to cultivate a compassionate educational landscape that not only predicts outcomes but also walks alongside learners and supports them on their unique academic journey.